\documentclass[final,3p]{elsarticle}

\usepackage{lineno,hyperref}
\modulolinenumbers[5]

\biboptions{authoryear}

\usepackage{natbib}

\usepackage[cmex10]{amsmath}
\usepackage{array}
\usepackage[caption=false,font=normalsize,labelfont=sf,textfont=sf]{subfig}
\usepackage{fixltx2e}
\usepackage{enumitem}
\usepackage{multirow}
\usepackage{amssymb}
\usepackage{amsthm}
\usepackage{pifont}
\usepackage{algorithm}
\usepackage[noend]{algpseudocode}
\usepackage{bm}
\usepackage{amsmath}

\usepackage{url}

\usepackage{booktabs}

\graphicspath{figs/}

\usepackage[draft,markup=none,commentmarkup=footnote]{changes}
\usepackage{color}

\usepackage{tikz}
\usepackage{pgfplots}
\usetikzlibrary{patterns}
\usepackage{csvsimple}
\usepackage{pgfplotstable}

\pgfplotsset{every tick label/.append style={font=\tiny}}
\pgfplotsset{
	box plot width/.initial=4em,
	box plot/.style={
		/pgfplots/.cd,
		black,
		only marks,
		mark=-,
		mark size=\pgfkeysvalueof{/pgfplots/box plot width},
		/pgfplots/error bars/.cd,
		y dir=plus,
		y explicit,
	},
	box plot box/.style={
		/pgfplots/error bars/draw error bar/.code 2 args={%
			\draw [line width=0.20mm]  ##1 -- ++(\pgfkeysvalueof{/pgfplots/box plot width},0pt) |- ##2 -- ++(-\pgfkeysvalueof{/pgfplots/box plot width},0pt) |- ##1 -- cycle;
		},
		/pgfplots/table/.cd,
		y index=2,
		y error expr={\thisrowno{3}-\thisrowno{2}},
		/pgfplots/box plot
	},
	box plot top whisker/.style={
		/pgfplots/error bars/draw error bar/.code 2 args={%
			\pgfkeysgetvalue{/pgfplots/error bars/error mark}%
			{\pgfplotserrorbarsmark}%
			\pgfkeysgetvalue{/pgfplots/error bars/error mark options}%
			{\pgfplotserrorbarsmarkopts}%
			\path ##1 -- ##2;
		},
		/pgfplots/table/.cd,
		y index=4,
		y error expr={\thisrowno{2}-\thisrowno{4}},
		/pgfplots/box plot
	},
	box plot bottom whisker/.style={
		/pgfplots/error bars/draw error bar/.code 2 args={%
			\pgfkeysgetvalue{/pgfplots/error bars/error mark}%
			{\pgfplotserrorbarsmark}%
			\pgfkeysgetvalue{/pgfplots/error bars/error mark options}%
			{\pgfplotserrorbarsmarkopts}%
			\path ##1 -- ##2;
		},
		/pgfplots/table/.cd,
		y index=5,
		y error expr={\thisrowno{3}-\thisrowno{5}},
		/pgfplots/box plot
	},
	box plot median/.style={
		/pgfplots/box plot
	}
}
\pgfplotsset{yticklabel style={text width=1.0em,align=right}}
\pgfplotsset{compat=1.12}

\definecolor{carrotorange}{rgb}{0.93, 0.57, 0.13}
\definecolor{darkcyan}{rgb}{0.0, 0.55, 0.55}
\definecolor{frenchblue}{rgb}{0.0, 0.45, 0.73}
\definecolor{brickred}{rgb}{0.8, 0.25, 0.33}
\definecolor{skymagenta}{rgb}{0.81, 0.44, 0.69}

\pgfplotstableread[col sep = comma]{plot_data/feat_sel.csv}\FeatSel
\pgfplotstableread[col sep = comma]{plot_data/feat_sel_scale.csv}\FeatSelScale
\pgfplotstableread[col sep = comma]{plot_data/MFGP_results_scale.csv}\MFGPScale
\pgfplotstableread[col sep = comma]{plot_data/MFGP_results_cell.csv}\MFGPCell
\pgfplotstableread[col sep = comma]{plot_data/pca_scale_high.csv}\PCAScaleHigh
\pgfplotstableread[col sep = comma]{plot_data/pca_scale_low.csv}\PCAScaleLow
\pgfplotstableread[col sep = comma]{plot_data/pca_cell_high.csv}\PCACellHigh
\pgfplotstableread[col sep = comma]{plot_data/pca_cell_low.csv}\PCACellLow

\begin{document}

\begin{frontmatter}
	\title{Multi-fidelity Gaussian Process for Biomanufacturing Process Modeling  with Small Data}
	
	\author[label1]{Yuan~Sun}
	\ead{yuan.sun@unimelb.edu.au}
	\author[label1]{Winton~Nathan-Roberts}
	\ead{winton.nathanroberts@student.unimelb.edu.au}
	\author[label1,label2]{Tien Dung~Pham}
	\ead{tiendungp@student.unimelb.edu.au}
	\author[label2]{Ellen~Otte}
	\ead{ellen.otte@csl.com.au}
	\author[label1]{Uwe~Aickelin\corref{cor1}}
	\ead{uwe.aickelin@unimelb.edu.au}
	\cortext[cor1]{Corresponding author}
	\address[label1]{School of Computing and Information Systems, The University of Melbourne, Parkville, VIC, 3010, Australia}
	\address[label2]{CSL Innovation, Parkville, VIC, 3052, Australia}

\begin{abstract}
In biomanufacturing, developing an accurate model to simulate the complex dynamics of bioprocesses is an important yet challenging task. This is partially due to the uncertainty associated with bioprocesses, high data acquisition cost, and lack of data availability to learn complex relations in bioprocesses. To deal with these challenges, we propose to use a statistical machine learning approach, multi-fidelity Gaussian process, for process modelling in biomanufacturing. Gaussian process regression is a well-established technique based on probability theory which can naturally consider uncertainty in a dataset via Gaussian noise, and multi-fidelity techniques can make use of multiple sources of information with different levels of fidelity, thus suitable for bioprocess modeling with small data. We apply the multi-fidelity Gaussian process to solve two significant problems in biomanufacturing, bioreactor scale-up and knowledge transfer across cell lines, and demonstrate its efficacy on real-world datasets.
\end{abstract}

\begin{keyword}
	Machine learning, data analytics, bioprocess modeling, multi-fidelity, small data 
\end{keyword}

\end{frontmatter}

\section{Introduction}

In the era of digital revolution, computers have been increasingly used to digitize and automate manufacturing processes \citep{glassey1994artificial,graefe1999new,lasi2014industry,frank2019industry}. In biologics manufacturing, efforts have been made to improve bioprocesses via advanced data analytics such as artificial intelligence and machine learning \citet{udugama2020role,gargalo2020towards,gargalo2020towards2}. However, there is still a long way to go to achieve full automation or digital twins in biomanufacturing, due to the high uncertainty of bioprocesses that involve living organisms and large, complex molecules \citep{sokolov2021hybrid}. 

To achieve a digital twin, one of the main challenges is building an accurate simulation model to mimic the complex dynamics of the underlying biosystem. This is of vital importance to improving the efficiency of bioprocesses, so as to satisfy the increasing demands for bioproduct. Having an accurate simulation model can help identify the optimal operating conditions for process control and design the best feed strategy for fed-batch cell culture, which would otherwise rely on expensive wet lab experiments which are also limited by the relatively low throughput of cell culture technologies \citep{bradford2018dynamic,duran2020multivariate}. Existing bioprocess modeling techniques can be roughly classified into three categories: mechanistic, data-driven, and hybrid \citep{solle2017between,del2019comparison}.

\textit{Mechanistic} methods are physics‐based models (e.g. differential equations) developed based on time-dependent mass balances of participating components in the biosystem \citep{kyriakopoulos2018kinetic}. Example mechanistic methods include the unstructured model developed by \citet{jang2000unstructured} to simulate the production of monoclonal antibodies in batch and fed-batch culture and the system of equations developed by \citet{del2017kinetic} to simulate and predict biomass growth and lutein production. Developing kinetic models requires deep understanding of underlying process mechanisms and significant biological knowledge. The knowledge learned via kinetic methods can typically be transferred between bioprocesses with similar underlying physical laws. However, the production of recombinant protein in mammalian cells cannot be fully described mechanistically based on our current biological knowledge and ability to measure cellular processes, so all mechanistic models of such processes involve assumptions that may impact on their usefulness. 
 
In contrast, \textit{data-driven} methods are statistical or machine learning models, that aim to automatically learn the underlying process mechanisms from experimental data \citep{glassey1994artificial}. Typical examples include the reinforcement learning approach proposed by \citet{petsagkourakis2020reinforcement} for batch bioprocess optimization, the Artificial Neural Network (ANN) presented by \citet{garcia2016artificial} to predict the growth of the microalga \textit{Karlodinium veneficum}, and the ANN used in \citep{del2017efficient} to model the rate of change of the dynamic biosystem. Data-driven methods are simple and easy to develop but typically require a large amount of high-quality data to train. In addition, predictions from data-driven models may be unreliable due to their black-box nature, and they have limited use for conditions outside the training dataset. 

\textit{Hybrid} methods, which combine the merits of both mechanistic and data-driven methods, have gained growing interest in recent years \citep{narayanan2019new,tsopanoglou2021moving,sokolov2021hybrid,merkelbach2022hybridml}. In general, hybrid methods make use of the mechanistic framework to improve model robustness while using data-driven techniques to improve model accuracy \citep{narayanan2019new}. For example, data-driven models can be used to estimate unknown parts of mechanistic models or to reduce the errors made by mechanistic models. Mechanistic models can be used to generate a large amount of data, which can then be used to improve the training of data-driven models \citep{solle2017between}.

Although many efforts have been made, the existing techniques are still insufficient to accurately capture the complex dynamics of bioprocesses \citep{sokolov2021hybrid}. Due to the biological variance of living cells \citep{fraser2001biological} and calibration or measurement errors, repeated wet lab experiments under the same conditions may lead to different system dynamics and product yield. Such a high level of uncertainty in bioprocesses makes system dynamics hard to predict, posing a significant challenge for existing techniques \citep{papathanasiou2020engineering}. Another challenge for bioprocess modeling is that acquiring data from wet lab experiments is very expensive (in cost of reagents and operators, as well as time), and hence the amount of data available is typically insufficient to train an accurate and robust model.

To address these challenges, we propose to use a statistical machine learning approach, Multi-Fidelity Gausssian Process (MFGP) \citep{Kennedy2000}, for bioprocess modeling. Gaussian Process (GP) is a data-driven approach that automatically learns a mapping from an input vector to an output, therefore not requiring deep knowledge of the underlying process mechanisms to build the model \citep{o1978curve,di2008biomass,bradford2018dynamic,deringer2021gaussian,petsagkourakis2021safe,del2021real}. GP can consider the uncertainty naturally inherent in a bioprocess as Gaussian noise and provide an uncertainty estimate along with the prediction. Multi-fidelity GP is a more advanced learning model that can make use of multiple sources of information with different levels of fidelity to build a prediction model \citep{Kennedy2000,Gratiet2014,Perdikaris2017, Peherstorfer2018}. Hence, it is particularly suitable for bioprocess modelling, for which the amount of high-fidelity data is typically small.

We apply the MFGP approach to model bioprocesses in which the amount of data is small, and demonstrate the efficacy of MFGP using two case studies: (1) bioreactor scale-up and (2) knowledge transfer across cell lines. For bioreactor scale-up, we use data collected from smaller-scale and larger-scale bioreactors as low-fidelity and high-fidelity data respectively. We show that using multiple sources of data can facilitate the development of a more robust and accurate model for larger-scale bioreactors than only using the data from larger-scale bioreactors. In the second case study, we treat the data collected from different cell lines as different levels of fidelity, and show that the knowledge in one cell line can be successfully transferred to another cell line via the MFGP approach. The contributions of this paper are summarized as follows:
\begin{enumerate}[label=(\alph*)] 
    \item We propose to use the MFGP approach for bioprocess modeling with a small amount of data. We show that the MFGP approach can potentially lead to an improved prediction model, especially when the amount of high-fidelity data is insufficient. 
    \item We apply the MFGP approach to solve two important tasks in bioprocess modeling, bioreactor scale-up and knowledge transfer across cell lines. The performance of the MFGP approach is evaluated on real-world datasets and compared against three other baseline methods. 
    \item We consider two typical MFGP methods, that can capture the linear correlation and nonlinear correlation between different levels of fidelity data respectively. The strength and weakness of different methods are thoroughly analysed on the two bioprocess modeling tasks.
\end{enumerate}

\section{Multi-fidelity Gaussian Process}

\subsection{Gaussian Process}
A GP is a stochastic process consisting of random variables $\{ f(x)\ |\ x\in S \}$ indexed by the index set $S=\{x\ |\ x\in \mathbb{R}^d\}$ such that any finite subset indexed by $\{x_1,\ \ldots,\ x_n\}\subseteq S$ is multivariate Gaussian distributed with mean $[m(x_1), \ldots, m(x_n)]^\top$ and covariance $k(x_i,x_j)=E\big{(}(f(x_i)-m(x_i))(f(x_j)-m(x_j))\big{)}$ where $m(x)=E(f(x))$ \citep{Rasmussen2018}:
\begin{equation}
f \sim GP(m(x),\ k(x_i,x_j)).
\end{equation}

Gaussian processes can be applied to regression analysis where $f$ represents the latent function to fit to observations $y(x_i)$ that are related through the observation model $y(x_i) = f(x_i) + \epsilon$ with $\epsilon \sim iid\ N(0,\sigma^2)$ representing measurement noise. The kernel function $k(x_i,x_j)$ is expressed in terms of learnable hyperparameters typically under a stationarity assumption, and encodes the general shape of the function. A common choice of kernel function is the radial basis function (RBF) kernel with length scale $\lambda$ as a hyperparameter and is defined as:
\begin{equation}
    k(x_i,x_j) = \exp \frac{-||x_i-x_j||^2}{2\lambda^2}.
\end{equation}

Without loss of generality it can be assumed that $m(x)=0$ which has a negligible effect on the predictive distribution particularly when the training data is standardised. The predictive distribution $f(x_*) | D, \theta, x_*$ of the latent function conditioned on test input $x_*$, training data $D$ and hyperparameters $\theta$ is Gaussian distributed with mean $E(f(x_*))$ and variance $V(f(x_*))$ given by:
\begin{align}
E(f(x_*)) &= k_*^T(K+\sigma I)^{-1}y,\\
V(f(x_*)) &= k(x_*, x_*) - k_*^T(K+\sigma I)^{-1}k_*,
\end{align}
where $k_*$ is a column vector of covariances between $f(x_*)$ and the training observations, and $K$ is a kernel matrix with entries $K_{ij} = k(x_i, x_j)$. An example of the predictive distribution of GP is shown in Figure~\ref{fig:illustration}.

Model selection can be done by choosing a hyperparameter setting $\hat \theta$ that maximises the marginal log likelihood (MLL) $p(Y|\theta, X)$, which is the log probability of the observed response $Y$ being a realisation of the model given the corresponding input $X$ in a procedure known as maximum likelihood type-II (ML-II) estimation as a frequentist approximation to a fully Bayesian treatment:
\begin{align}
\log p(Y|\theta, X) &= -\frac{1}{2}(y^T(K+\sigma I)^{-1}y+ \log |K+\sigma I| + n \log 2\pi),\\
\hat\theta &= \text{argmax}_{\theta\in \mathbb{R}} \log p(Y|X,\ \theta).
\end{align}
The MLL penalises model misfit $y^T(K+\sigma I)^{-1}y$ and model complexity $\log |K+\sigma I|$ which trades off between goodness of fit and overfitting respectively. 

\subsection{Multi-fidelity Gaussian Process}

\begin{figure}[!t]
    \centering
    \resizebox{\textwidth}{!}{
    \includegraphics[scale=0.5]{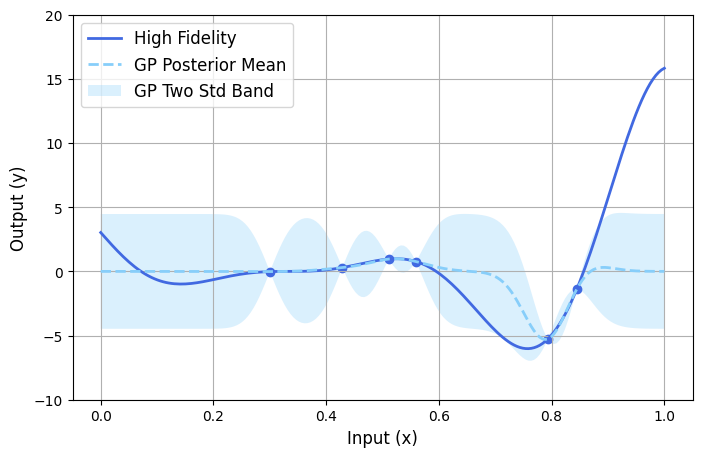}
    \includegraphics[scale=0.5]{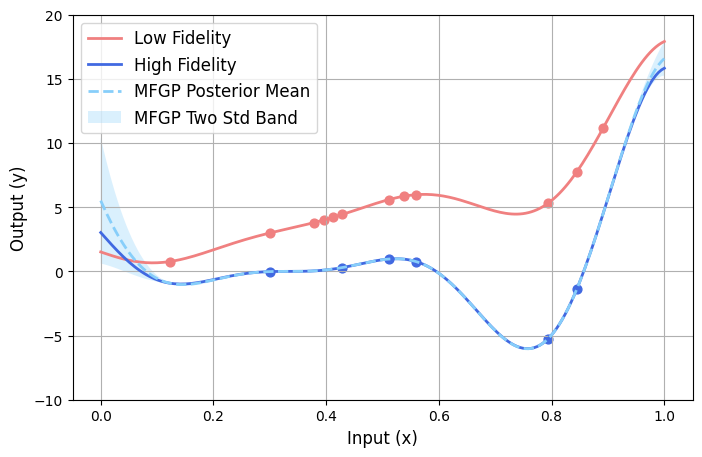}
    }
    \caption{An illustration of the GP and MFGP models. In the left figure, only six data points from the high fidelity function are available to train a GP model, of which the posterior mean and two standard deviation band are shown. In the right figure, in addition to the six high-fidelity data points, twelve data points from the low fidelity function are available. An MFGP model is trained using both the low-fidelity and high-fidelity data, with the posterior mean and two standard deviation band shown.}
    \label{fig:illustration}
\end{figure}

It is assumed fidelity levels $\{X_i, Y_i\}_{1\leq i \leq L}$ are available in the training dataset that describe the highest fidelity level $Y_L$ with varying degrees of fidelity, where $X_i$ is the input that generated the response $Y_i$. Fidelity level $i<j$ implies observations of fidelity $i$ are a lower fidelity estimate of $Y_L$ compared to those of level $j$, forming a hierarchy. It is assumed observations $y_i \in Y_t$ for corresponding input $x_i\in X_t$ are related to an underlying latent function $f_i$ through the observation model $y_i = f_i(x_i) + \epsilon$ with $\epsilon \sim iid\ N(0, \sigma)$ representing obscuring noise. The fidelity levels can be related through an autoregressive function $f$ that expresses level $t$ in terms of level $t-1$ for $t\geq 1$ which exploits the correlation between levels for their prediction:
\begin{equation}\label{eq:0}
\begin{aligned}
f_t(x) &= f(f_{t-1}(x), x).
\end{aligned}
\end{equation}
An example of the MFGP approach is shown in Figure~\ref{fig:illustration}.

\subsubsection{Linear Autoregressive Gaussian Process (LARGP)}

A family of multi-fidelity models known as multi-fidelity Gaussian process models exist in literature which differ by the choice of $f$ in which Gaussian processes are assigned as priors to equation terms resulting in a non-parametric and uncertainty propagating Bayesian statistical model that is ideal for small-data regimes. The seminal work of \citet{Kennedy2000} assumes fidelity level $t$ is some linear model of the lower fidelity $t-1$ in terms of a hyperparameter scaling factor $\rho_t$ plus some error correction $\delta_t(x)$ that is assigned a GP prior:
\begin{equation}\label{eq:1}
\begin{aligned}
f_1(x)& = \delta_1(x),\\
f_t(x) &= \rho_t f_{t-1}(x) + \delta_t(x), \text{ for $t\geq 2$}.
\end{aligned}
\end{equation}
By assuming independence between $\delta_t(x)$ and lower fidelity levels $(f_{m}(x))_{m < t}$, a closed form solution can be derived. Due to this assumption, given $f_{t-1}(x)$ nothing more can be learned from lower fidelity levels about $f_t(x)$ (Markov property).

\citet{Gratiet2014} propose a computationally efficient recursive method for computing the posterior, involving sequentially fitting independent multi-fidelity models to fidelity level pairs $(t-1,t)$ beginning with the lowest fidelity level under a nested data assumption $X_i \subseteq X_j$ for fidelity levels $i<j$. The key difference in their solution is to replace $f_{t-1}(x)$ in Equation~\eqref{eq:1} with its posterior $f_{t-1}^*(x)$ conditioned on all lower fidelity levels. As they prove, the resulting predictive distribution  is identical to the coupled model proposed by \citet{Kennedy2000}. In the model each fidelity level can be modelled with GP regression over the observations for the corresponding fidelity level and the posterior of the lower fidelity level $f_{t-1}^*(x)$, where the predictive distribution for test input $x_*$ is Gaussian distributed with mean $E(f_t (x_*))$ and variance $V(f_t (x_*))$ given by:
\begin{align}\label{eq:3}
E(f_t (x_*)) &= \rho_t E(f_{t-1} (x_*))+\mu_t+ k_*^T(K_t+\sigma_t I)^{-1}\large{(}\normalsize y_t -\rho_tE(f_{t-1} (x_*))-\mu_t\large{)},\\
V(f_t (x_*)) &= \rho_t^2  V(f_{t-1} (x_*))+k(x_*, x_*)- k_*^T(K_t+\sigma_t I)^{-1}k_*,
\end{align}
where $\mu_t$ is the mean of $\delta_t$, $K_t$ is the kernel matrix of the response $f_t(x)$ defined in terms of kernel hyper parameters $\theta_t$, and $k_*$ is a column vector of covariances between the test point and training points. The time complexity is significantly reduced to $O(\sum_i n_i^3)$ due to the inversion of square matrices of size $n_i$ for the sequential inference of each fidelity level $i$ having $n_i$ observations. In the following sections, this recursive model is used in an application to bioprocess modelling as introduced denoted MFGP.

\subsubsection{Nonlinear Autoregressive Gaussian Process (NARGP)}
\citet{Perdikaris2017} replace the restrictive linear function that relates adjacent fidelities in Equation~\eqref{eq:1} with an expressive GP $F_t$ that can capture nonlinear relationships:
\begin{equation}\label{eq:4}
\begin{aligned}
f_1(x) &= \delta_{1}(x), \\
f_t(x) &= F_{t}(f_{t-1}(x),x), \; \text{ for $t \geq 2$},
\end{aligned}
\end{equation}
where $\delta_1$ is a GP over the input $x$. Since $F_t(x)$ is the composition of GPs known as a Deep Gaussian Process (DGP), the posterior is no longer Gaussian \citep{Damianou2012}. To arrive at a solution, \citet{Perdikaris2017} replace $f_{t-1}(x)$ with its posterior $f_{t-1}^*(x)$ which assumes $f_t$ is independent of all fidelity levels given $f_{t-1}$. The chosen kernel is:
\begin{equation}\label{eq:5}
\begin{aligned}
K_1 &= K_b^1(x_i, x_j), \\
K_t &= K_d^t(x_i, x_j)K_f^t(f_{t-1}(x_i),f_{t-1}(x_j)) + K_b^t(x_i, x_j), \; \text{ for $t \geq 2$}, 
\end{aligned}
\end{equation}
where RBF kernels $K_f$ and $K_d$ describe the interaction between the input $x$ and the lower fidelity $f_{t-1}$, and RBF kernel $K_b$ represents the covariance of the bias. The time complexity is similar to the recursive solution to LARGP proposed by \citet{Gratiet2014}.







\section{Bioprocess Scale-up}\label{sec::MFGP4scale-up}

Bioprocess scale-up plays an important role in biomanufacturing, as it underlies the transition from laboratory-scale early process development to a large-scale production environment \citep{lindskog2018upstream,richelle2020analysis}. This often involves a change in the fermentation conditions, ranging from the geometries of the bioreactors to the operating speed or flow rates of substrate supply and additions of reagents. Although small-scale data is easier to acquire, with lower experimentation cost and faster turnaround time, such data may not accurately represent the production-scale environment. As bioprocesses involve living cells which adapt and react differently to their environment, scale-up can significantly change the cell culture behaviour, resulting in changes in productivity, yield and product quality. Consequently, the bioprocesses do not simply scale according to volumetric changes, but may require a modified model to represent the new biodynamics at the larger scale \citep{xia2015advances}. These differences in data fidelity pose a challenge in using lab-scale results as training data to model bioprocesses across different scales. 

Existing studies have used multivariate data analysis for bioprocess scale-up \citep{mercier2013multivariate,tescione2015application}. Such a simple statistical approach is often inadequate to address the challenges inherent in bioprocess modeling. In this paper, we propose to use a more advanced statistical machine learning approach, MFGP, which naturally treats the data from different scales with different levels of fidelity. The MFGP approach has the potential to automatically learn complex relationships between the data generated from different scales, leading to a more robust machine learning model, even if the amount of high-fidelity data is small. Hence, MFGP is very suitable for bioprocess scale-up, as the amount of available data from large-scale bioreactors is typically small due to high data acquisition cost. 

\subsection{Dataset}

\begin{figure}[!t]
\centering
\begin{tikzpicture}
	\begin{axis} [box plot width=0.20em, xlabel = \scriptsize First Principal Component of Feed Medium,  ylabel = \scriptsize Normalized Product Concentration,    height=0.50\textwidth,width=0.95\textwidth, grid style={line width=.1pt, draw=gray!10},major grid style={line width=.2pt,draw=gray!30}, xmajorgrids=true, ymajorgrids=true,  major tick length=0.05cm, minor tick length=0.0cm, legend style={at={(0.65,0.80)},anchor=west,font=\scriptsize,draw=none}]
 	\addplot[only marks, color=brickred, mark=*, mark size = 2.0] table[x index=0, y index=1] {\PCAScaleHigh}; \addlegendentry{\scriptsize Data from 5L Bioreactors}
 	\addplot[only marks, color=frenchblue, mark=diamond*, mark size = 2.0] table[x index=0, y index=1] {\PCAScaleLow}; \addlegendentry{\scriptsize Data from 0.25L Bioreactors};
 	\end{axis}
\end{tikzpicture}
\caption{The distribution of data points from bioreactors with different scales. The x axis is the first principal component of the feed medium and the y axis is the normalized recombinant protein concentration.}
\label{fig:pca_scale}
\end{figure}
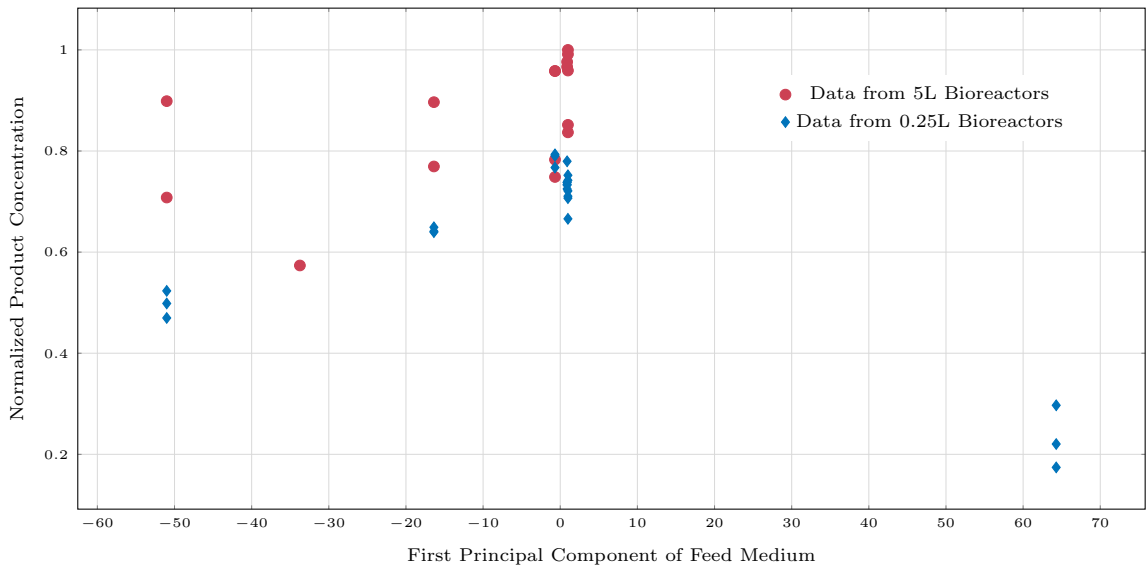

As a case study, we use the data generated from 40 fed-batch bioreactors with a culture period of 14 days. This includes 24 data points from Ambr® 250 bioreactors at the 0.25L scale and 16 data points from bioreactors at the 5L scale. The same CHO cell line was fermented across all bioreactors under different feeding strategies. For each strategy, we recorded the concentration of twenty amino acid components in the feed medium. The target variable is the end-point (at day 14) measurement of recombinant protein concentration, which is normalized into a range of 0 to 1. The distribution of the 40 data points is shown in Figure~\ref{fig:pca_scale} via Principal Component Analysis (PCA). Our goal is to model the effects of feed medium on the recombinant protein concentration, which can further be used to optimise the feeding strategy at different scales. We build a MFGP model for the prediction task for 5L bioreactors, treating the 0.25L mini-bioreactor data as the low-fidelity level and the 5L-scale bioreactor data as the high-fidelity level.

\subsection{Feature Selection}\label{subsec::fs_scale}

In this dataset, the number of high-fidelity data points is less than the number of features (i.e. amino acids). Using all features to train a machine learning model likely leads to overfitting. We therefore identify a subset of important features that are mostly relevant to the target (i.e. recombinant protein concentration), and use the selected subset of features to train GP models. To do so, we use a widely-used information theoretic feature selection method, Minimum-Redundancy Maximum-Relevance (MRMR) \citep{peng2005feature}, which uses (conditional) mutual information to measure the relevance between features and the target. As the features and target in this dataset are continuous variables, we use the Minimum Description Length method \citep{fayyad1993multi} to evenly divide the continuous values into five bins, following \citep{sun2020revisiting}. Note that the discretized dataset is only used for the feature selection procedure, whilst the original continuous dataset is used in the subsequent prediction tasks. 

\begin{figure}[!t]
\centering
\begin{tikzpicture}
	\begin{axis} [box plot width=0.20em, xlabel = \scriptsize Number of features selected ($N_f$),  ylabel = \scriptsize RMSE of LARGP,    height=0.50\textwidth,width=0.95\textwidth, grid style={line width=.1pt, draw=gray!10},major grid style={line width=.2pt,draw=gray!30}, xmajorgrids=true, ymajorgrids=true,  major tick length=0.05cm, minor tick length=0.0cm, legend style={at={(0.05,0.10)},anchor=west,font=\scriptsize,draw=none}]
 	\addplot[color=brickred, mark=*, mark size = 2.0, line width=0.50mm, dashed] table[x index=0, y index=1] {\FeatSelScale};
 		\end{axis}
\end{tikzpicture}
\caption{The RMSE generated by LARGP when using different number of features selected by MRMR for bioprocess scale-up.}
\label{fig:feature_selection_scale}
\end{figure}
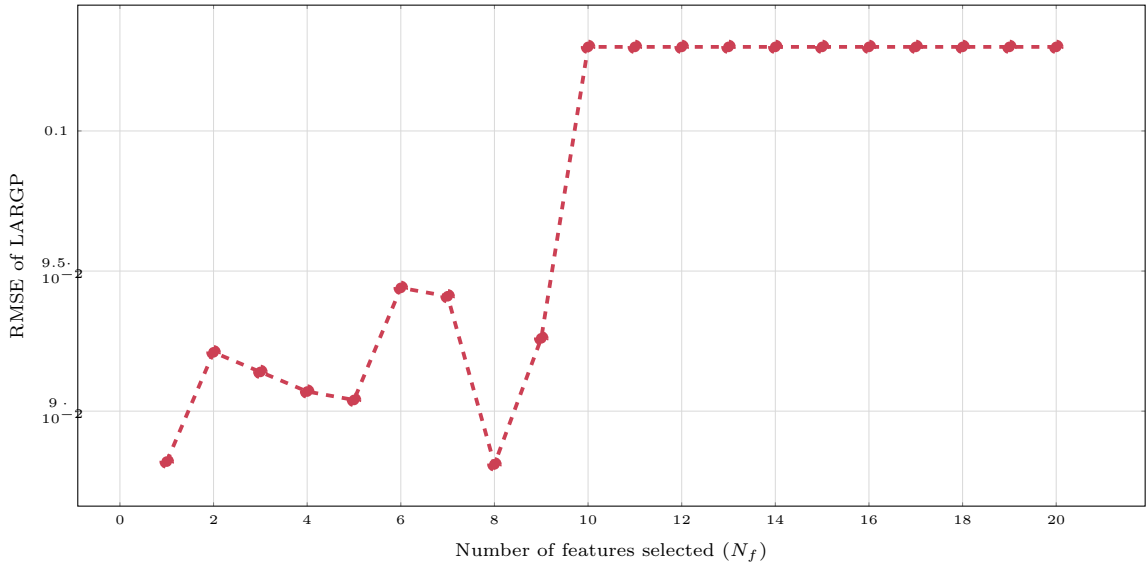

The MRMR method essentially provides a ranking of the features based on their relevance to the target. To determine an appropriate subset size, we vary the number of selected features ($N_f$) from 1 to 20, and train a separate LARGP model using the selected features. For each model, fifteen high-fidelity data points are used in training and one data point is left for testing. The root mean square errors (RMSE) averaged across 30 independent runs (with different random seeds) are presented in Figure~\ref{fig:feature_selection_scale}. The results show that feature selection is beneficial in the sense that using a subset of selected features to train the LARGP model can result in a smaller RMSE than using all the features (i.e. $N_f=20$). The smallest RMSE is achieved when using one or eight features. In the following, we will select eight features to train all the models. 

\subsection{Comparison between MFGP Models}\label{subsec::mgfp_comp_scale}

We consider two MFGP models, LARGP and NARGP, for the prediction task. It is important to note that the MFGP models require the nested data structure of the training data, which means for every high-fidelity data point, there must exist a corresponding low-fidelity data point with the same the feature vector (feed medium). However, this assumption might be violated as can be seen in Figure~\ref{fig:pca_scale}. To address this, we train a GP model on the low-fidelity data and use the trained model to sample new data points to satisfy the nested data assumption. 

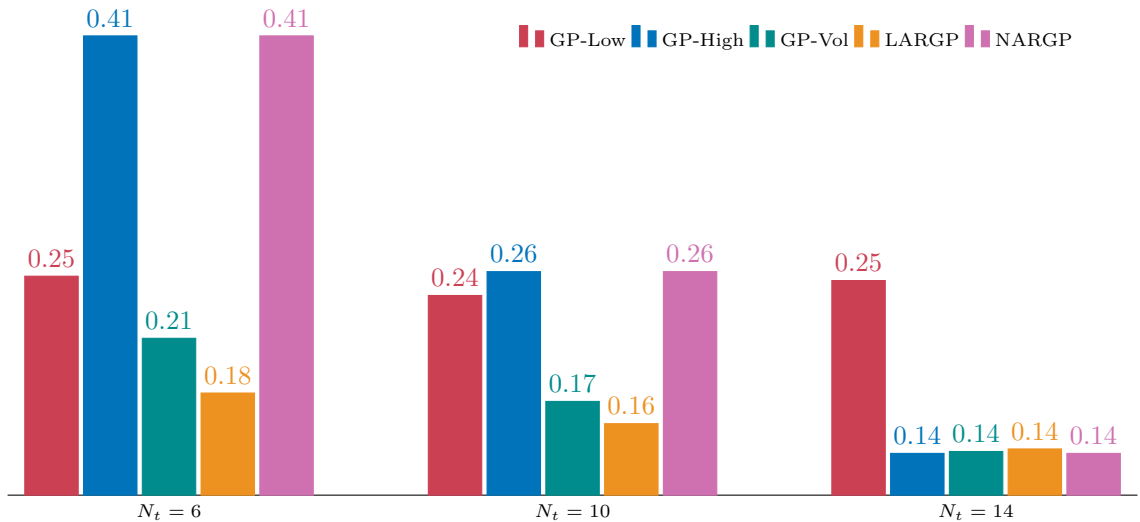
\begin{figure*}[!t]
	\centering
	\begin{tikzpicture}
	\begin{axis}[ybar, symbolic x coords={6, 10, 14}, xtick=data,  ytick=\empty,  axis y line = none, axis x line* = bottom, enlarge x limits = 0.2, nodes near coords={\pgfmathprintnumber[fixed zerofill, precision=2]{\pgfplotspointmeta}},  height=0.50\textwidth,width=1.0\textwidth, bar width=20pt, xticklabels={\scriptsize $N_t=6$, \scriptsize $N_t=10$,\scriptsize $N_t=14$}, major tick length=0.0cm, minor tick length=0.0cm, legend style={at={(0.7,0.95)},anchor=north,legend columns=0, font=\scriptsize,draw=none}]
	\addplot[color=brickred, fill=brickred] table[x=Category, y=GP-Low] {\MFGPScale};\addlegendentry{\scriptsize GP-Low}
	\addplot[color=frenchblue, fill=frenchblue] table[x=Category, y=GP-high] {\MFGPScale};\addlegendentry{\scriptsize GP-High}
	\addplot[color=darkcyan, fill=darkcyan] table[x=Category, y=GP-vol] {\MFGPScale};\addlegendentry{\scriptsize GP-Vol}
	\addplot[color=carrotorange, fill=carrotorange] table[x=Category, y=AR1] {\MFGPScale};\addlegendentry{\scriptsize LARGP}
	\addplot[color=skymagenta, fill=skymagenta] table[x=Category, y=NARGP] {\MFGPScale};\addlegendentry{\scriptsize NARGP}
	\end{axis}
	\end{tikzpicture}
    \caption{The RMSE generated by the machine learning models for bioprocess scale-up. $N_t$ is the number of high-fidelity data points used in training with the remaining data points used for testing. The RMSE is averaged across 30 independent runs with different random seeds.}
    \label{fig:mfgp-scale}
\end{figure*}

The results generated by LARGP and NARGP are presented in Figure~\ref{fig:mfgp-scale}. The linear autoregressive model, LARGP, performs much better than the non-linear autoregressive model, NARGP, on this dataset. The RMSE generated by the NARGP model is the same as that of GP-High, which is a GP model trained on high-fidelity data only. This indicates that the NARGP model fails to learn any correlation between high-fidelity and low-fidelity data, possibly due to the high uncertainty inherent in the dataset as can be seen in Figure~\ref{fig:pca_scale}. Hence, we will mainly consider LARGP for bioprocess scale-up in the following.  

\subsection{Comparison to Baselines}

We compare the LARGP model to three baselines: 1) GP-Low, trained on low-fidelity data only, 2) GP-High, trained on high-fidelity data only, and 3) GP-Vol, trained on both low-fidelity and high-fidelity data with the volume of bioreactors as an additional feature. The RMSE generated by each method is shown in Figure~\ref{fig:mfgp-scale}. The LARGP model generally achieves the lowest RMSE for the task of bioprocess scale-up compared to all other methods. This demonstrates its efficacy in learning the correlation between data from different fidelities and smart use of multiple sources of information to build a better predictive model. The GP-Vol is essentially another linear autoregressive model, also learning the correlation between low-fidelity and high-fidelity data. However, it focuses on predicting for both low-fidelity and high-fidelity data, in contrast to LARGP which aims to predict for high-fidelity data only. 

\subsection{Effects of the Number of Training Data Points}
There are totally 16 high-fidelity data points in our dataset. Here, we vary the number of high-fidelity data points used to train the GP models ($N_t = $ 6, 10, or 14), and the results are shown in Figure~\ref{fig:mfgp-scale}. As the number of high-fidelity training data points increases, the RMSE generated by all models (except GP-Low) decreases. The GP-Low model is trained on low-fidelity data only, thus its performance is not improved when increasing the number of high-fidelity training data points. When $N_t = 6$, the RMSE generated by GP-Low is much smaller than that of GP-High, indicating that the low-fidelity data is related to the high-fidelity data. As the number of high-fidelity training data points increases, the difference between LARGP, GP-High, and GP-Vol decreases. This suggests that the methodology of utilising multiple sources of data to build a predictive model is more significant when the number of high-fidelity training data points is insufficient. In other words, if the number of high-fidelity training data points is sufficient, we can simply use the GP-High model. 
\section{Knowledge Transfer across Cell Lines}\label{sec::MFGP4cell-lines}

Similar to scaleability, cell line is another crucial process parameter that  alters  cell culture behaviour. A cell line is a population of transformed cells that can be divided indefinitely while maintaining stability of certain phenotypes and functions \citep{lindskog2018host}. Different cell lines are developed for each biological product through a process of cell line development.   A desirable cell line should be highly adaptive to the fermentation conditions, such that it can retain a stably high growth rate and productivity \citep{CASTAN2018131}. Knowledge transfer across cell lines is highly desirable when developing a new bioproduct, as it can potentially harness the historical experimental data of different cell lines to reduce the high cost of generating new data \citet{hutter2021knowledge,rogers2022transfer}. However, the underlying relationship between different cell lines is often complex and difficult for data-driven methods to learn. Here, we propose that the MFGP approach is a powerful technique that can automatically learn the relationship between different cell lines from data and provides a natural approach for knowledge transfer across cell lines. 

\subsection{Dataset}
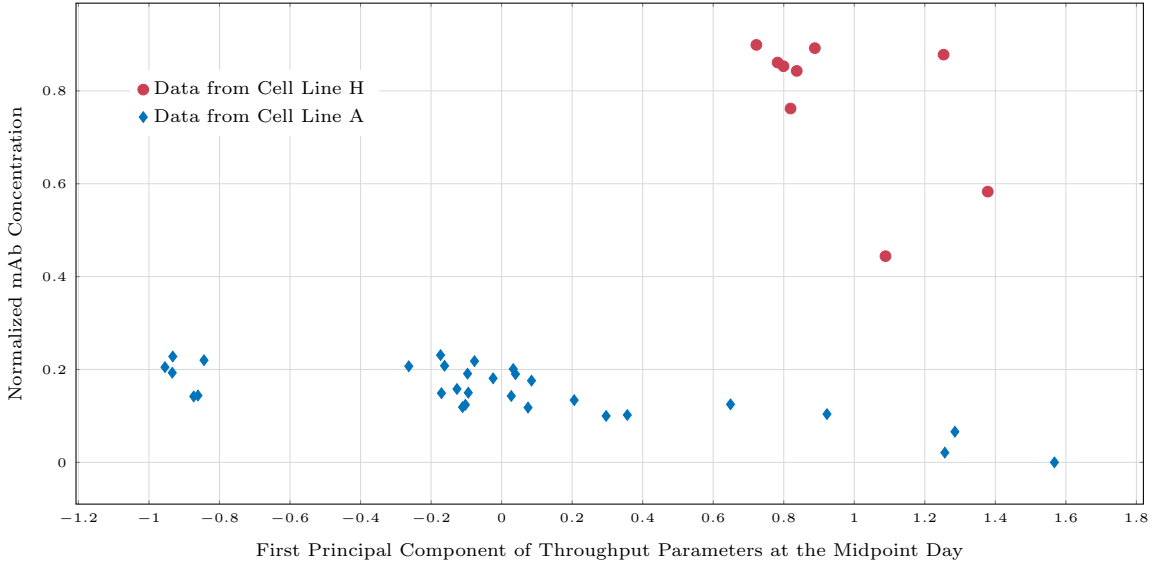
\begin{figure}[!t]
\centering
\begin{tikzpicture}
	\begin{axis} [box plot width=0.20em, xlabel = \scriptsize First Principal Component of Throughput Parameters at the Midpoint Day,  ylabel = \scriptsize Normalized mAb Concentration,    height=0.50\textwidth,width=0.95\textwidth, grid style={line width=.1pt, draw=gray!10},major grid style={line width=.2pt,draw=gray!30}, xmajorgrids=true, ymajorgrids=true,  major tick length=0.05cm, minor tick length=0.0cm, legend style={at={(0.05,0.80)},anchor=west,font=\scriptsize,draw=none}]
 	\addplot[only marks, color=brickred, mark=*, mark size = 2.0] table[x index=0, y index=1] {\PCACellHigh}; \addlegendentry{\scriptsize Data from Cell Line H}
 	\addplot[only marks, color=frenchblue, mark=diamond*, mark size = 2.0] table[x index=0, y index=1] {\PCACellLow}; \addlegendentry{\scriptsize Data from Cell Line A};
 	\end{axis}
\end{tikzpicture}
\caption{The distribution of data points from different cell lines. The x axis is the first principal component of the throughput parameters at the midpoint day and the y axis is the normalized mAb concentration.}
\label{fig:pca_cell}
\end{figure}

As a case study, we focus on Chinese Hamster Ovary (CHO) cell lines, which are primarily used in the biopharmaceutical manufacturing industry to produce recombinant monoclonal antibodies (mAb). Different recombinant protein products often require different CHO cell lines, for example, CHO-K1, CHO-RD, or CHO.1F8 \citep{ZHANG2013127}. For this case study, we use the dataset presented in \citep{gangadharan2021data}, which was extracted from AstraZeneca upstream process development and production databases. For each cell culture, the 22 throughput parameters (e.g., Glutamine and Glutamate) were recorded for 17 days. The task was to predict the mAb concentration at the endpoint day using the throughput parameters at the midpoint day \citep{gangadharan2021data}. The values of each parameter were normalized to the range between 0 and 1, and the categorical parameters were anonymised using alphabet letters. There are 30 data points from cell line A and 9 data points from cell line H, and the data distribution is visualised in Figure~\ref{fig:pca_cell} via PCA. Our aim is to build a MFGP model for knowledge transfer between cell lines A and H. Specially, we use the data from cell line A as low-fidelity data and that from cell line H as high-fidelity data. 

\subsection{Feature Selection}

\begin{figure}[!t]
\centering
\begin{tikzpicture}
	\begin{axis} [box plot width=0.20em, xlabel = \scriptsize Number of features selected ($N_f$),  ylabel = \scriptsize RMSE of LARGP, height=0.50\textwidth,width=0.95\textwidth, grid style={line width=.1pt, draw=gray!10}, major grid style={line width=.2pt,draw=gray!30}, xmajorgrids=true, ymajorgrids=true, major tick length=0.05cm, minor tick length=0.0cm, legend style={at={(0.05,0.10)},anchor=west,font=\scriptsize,draw=none}]
 	\addplot[color=brickred, mark=*, mark size = 2.0, line width=0.50mm, dashed] table[x index=0, y index=1] {\FeatSel};
 	\end{axis}
\end{tikzpicture}
\caption{The RMSE generated by LARGP when using different number of features selected by MRMR for knowledge transfer across cell lines.}
\label{fig:feature_selection}
\end{figure}
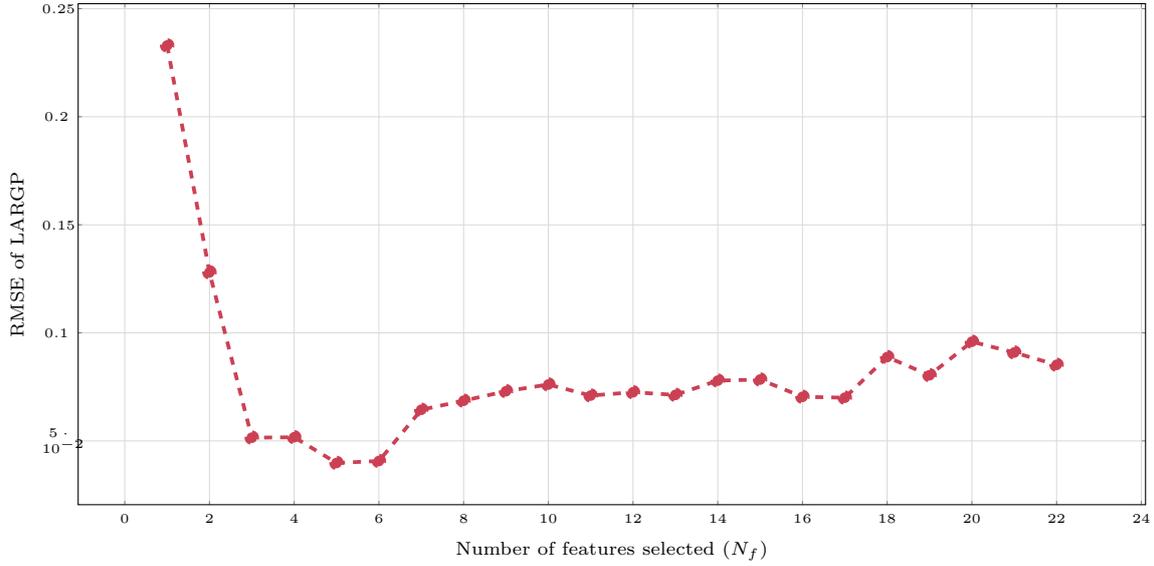

We first use the MRMR method to select a subset of relevant features for this prediction task, similar to that in Section~\ref{subsec::fs_scale}. The RMSE generated by LARGP when using different number of selected features is shown in Figure~\ref{fig:feature_selection}. We can observe that using a subset of selected features to train the LARGP model can lead to a smaller RMSE as compared to using all the features (i.e. $N_f=22$). The smallest RMSE is generated when using six features to train the LARGP model. However, as the number of high-fidelity data points in this dataset is only nine, using six features to train a machine learning model can still lead to overfitting, especially for the GP model trained only on the high-fidelity data. Hence in our subsequent study, we will select three features, for which the RMSE generated by LARGP is also low. 

\subsection{Comparison between MFGP Models}

\begin{figure*}[!t]
	\centering
	\begin{tikzpicture}
	\begin{axis}[ybar, symbolic x coords={3, 5, 7}, xtick=data,  ytick=\empty,  axis y line = none, axis x line* = bottom,  enlarge x limits = 0.2, nodes near coords,
    nodes near coords style={/pgf/number format/fixed, /pgf/number format/precision=3}, height=0.50\textwidth,width=1.0\textwidth, bar width=20pt, xticklabels={\scriptsize $N_t=3$, \scriptsize $N_t=5$,\scriptsize $N_t=7$}, major tick length=0.0cm, minor tick length=0.0cm, legend style={at={(0.40,1.05)},anchor=north,legend columns=0, font=\scriptsize,draw=none}]
    \addplot[color=brickred, fill=brickred] table[x=Category, y=GP-Low]{\MFGPCell};\addlegendentry{\scriptsize GP-Low}
	\addplot[color=frenchblue, fill=frenchblue] table[x=Category, y=GP-High] {\MFGPCell};\addlegendentry{\scriptsize GP-High}
	\addplot[color=darkcyan, fill=darkcyan] table[x=Category, y=GP-Cat] {\MFGPCell};\addlegendentry{\scriptsize GP-Cat}
	\addplot[color=carrotorange, fill=carrotorange] table[x=Category, y=AR1] {\MFGPCell};\addlegendentry{\scriptsize LARGP}	
	\addplot[color=skymagenta, fill=skymagenta] table[x=Category, y=NARGP] {\MFGPCell};\addlegendentry{\scriptsize NARGP}
	\end{axis}
	\end{tikzpicture}
    \caption{The RMSE generated by the machine learning models for knowledge transfer between cell lines. $N_t$ is the number of high-fidelity data points used in training with the remaining data points used for testing. The RMSE is averaged across 30 independent runs with different random seeds.}
    \label{fig:mfgp-cell}
\end{figure*}
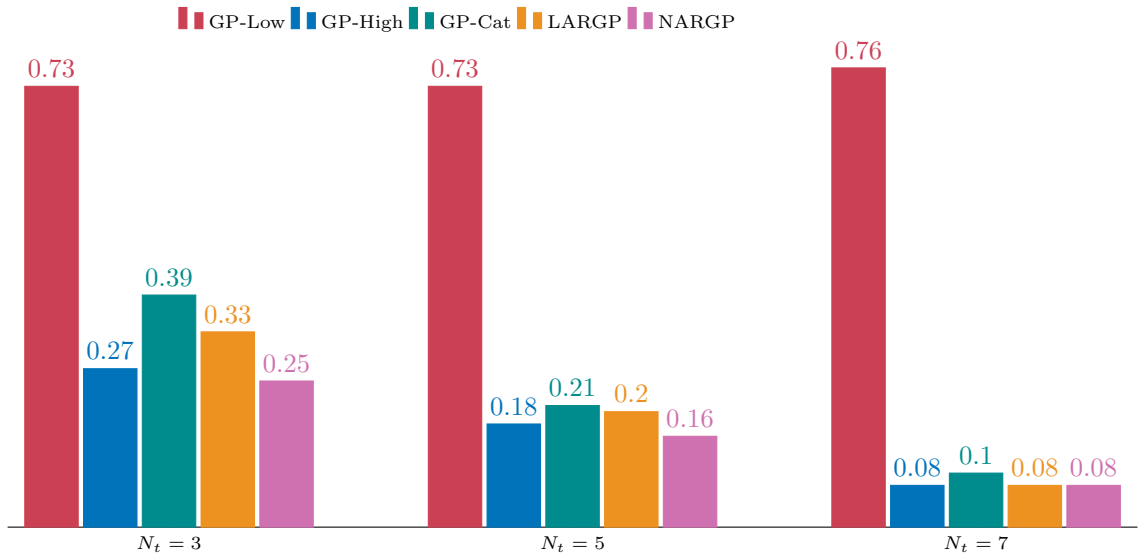

We compare the performance of LARGP and NARGP on the task of knowledge transfer across cell lines. The nested data assumption is satisfied by using the same method as in Section~\ref{subsec::mgfp_comp_scale}. The RMSEs generated by both models are presented in Figure~\ref{fig:mfgp-cell}. Interestingly, the non-linear autoregressive model (NARGP) performs better than the linear autoregressive model (LARGP) on this dataset, suggesting the correlation between the data from the cell lines is non-linear. Hence, we will mainly consider NARGP for knowledge transfer across cell lines in our subsequent study. 

\subsection{Comparison to Baselines}

We compare the NARGP model to three baselines: 1) GP-Low, trained on low-fidelity data only, 2) GP-High, trained on high-fidelity data only, and 3) GP-Cat, trained on both low-fidelity and high-fidelity data with the category of cell lines (i.e., 1 and 2) as an additional feature. Note that similar to LARGP, the GP-Cat model is equivalent to a linear autoregressive model that is able to learn any linear correlation between the low-fidelity and high-fidelity data. However, the GP-Cat model aims to minimise the prediction error for both low-fidelity and high-fidelity data, whilst the LARGP model aims to minimise the prediction error for high-fidelity data only. 

The RMSEs generated by all the models are presented in Figure~\ref{fig:mfgp-cell}. The GP-Low model performs much worse than other models, suggesting that the low-fidelity data is significantly different from high-fidelity data in this dataset and the knowledge about cell line A cannot be directly transferred to cell line H. Interestingly, the two linear autoregressive models, LARGP and GP-Cat, are generally outperformed by the GP-High model. This again indicates that the correlation between low-fidelity and high-fidelity data in this dataset is nonlinear, requiring a more sophisticated model such as NARGP. As expected, the NARGP model achieves the lowest RMSE for this prediction task as compared to all other models. 

\subsection{Effects of the Number of Training Data Points}

We investigate the effect of the number of training data points on the performance of the machine learning models. To do so, we vary the number of high-fidelity data points used in training ($N_t = $ 3, 5, or 7), with the remaining high-fidelity data points for testing. The results in Figure~\ref{fig:mfgp-cell} show that the performance of all models improves (except GP-Low) when more high-fidelity training data points are available. The performance of GP-Low does not improve as it is trained on low-fidelity data. The efficacy of the NARGP model is more significant when a smaller amount of high-fidelity training data is available.

\section{Conclusion}
In this article, we have proposed to use a statistical machine learning approach named Multi-Fidelity Gaussian Process (MFGP) for process modelling in biomanufacturing. We have shown that this MFGP technique provides a powerful tool to deal with the small data challenge in bioprocess modelling, by making use of multiple sources of data with different levels of fidelity. We have successfully applied MFGP for bioreactor scale-up and knowledge transfer across cell lines, representing two important problems in biomanufacturing. For bioreactor scale-up, we demonstrated that using the data collected from small bioreactors as low-fidelity data can yield an improved model for the prediction task for large bioreactors. Similarly, for knowledge transfer across cell lines, we treated the data generated from different cell lines as different levels of fidelity, and showed that the pattern underlying one cell line (i.e., the one treated as low-fidelity data) can be successfully transferred to another (i.e., the one treated as high-fidelity). We showed that the MFGP technique can achieve a higher accuracy for these two prediction tasks as compared to three baseline methods, especially when the amount of training data is small. 

We have considered two MFGP models in this paper, a linear autoregressive model (LARGP) and a nonlinear autoregressive model (NARGP). The results showed that 1) for bioprocess scale-up, the LARGP model performs better as the corresponding dataset is too noisy for the NARGP model to learn any complex correlation between data from different scales; and 2) for knowledge transfer across cell lines, the NARGP model performs better as the correlation between the cell lines is nonlinear in the dataset considered. In future work, it will be interesting to consider more MFGP models (e.g. \citep{cutajar2019deep}) and develop a method that can automatically select a well-performed MFGP model based on the characteristics of the dataset under consideration. This may be achieved by using algorithm selection techniques \citep{munoz2015algorithm} and instance space analysis \citep{andres2022bifidelity}. There may exist other opportunities of leveraging MFGP for bioprocess modeling. For example, we can use simulation data generated from a mechanistic model \citep{kyriakopoulos2018kinetic} as low-fidelity data and real data from bioreactors as high-fidelity data. This can potentially lead to a novel and effective hybrid model \citep{tsopanoglou2021moving,sokolov2021hybrid}, suitable for tackling the challenge of small data in bioprocess modeling. 





\section*{Acknowledgement}
This research was financially supported by the Victorian High Education State Investment Fund (VHESIF) Smarter, Faster, Biopharma and Food Manufacturing and CSL Innovation Ltd.


\bibliography{reference}

\begin{thebibliography}{51}
\expandafter\ifx\csname natexlab\endcsname\relax\def\natexlab#1{#1}\fi
\providecommand{\url}[1]{\texttt{#1}}
\providecommand{\href}[2]{#2}
\providecommand{\path}[1]{#1}
\providecommand{\DOIprefix}{doi:}
\providecommand{\ArXivprefix}{arXiv:}
\providecommand{\URLprefix}{URL: }
\providecommand{\Pubmedprefix}{pmid:}
\providecommand{\doi}[1]{\href{http://dx.doi.org/#1}{\path{#1}}}
\providecommand{\Pubmed}[1]{\href{pmid:#1}{\path{#1}}}
\providecommand{\bibinfo}[2]{#2}
\ifx\xfnm\relax \def\xfnm[#1]{\unskip,\space#1}\fi
\bibitem[{Andr{\'e}s-Thi{\'o} et~al.(2022)Andr{\'e}s-Thi{\'o}, Mu{\~n}oz \&
  Smith-Miles}]{andres2022bifidelity}
\bibinfo{author}{Andr{\'e}s-Thi{\'o}, N.}, \bibinfo{author}{Mu{\~n}oz, M.~A.},
  \& \bibinfo{author}{Smith-Miles, K.} (\bibinfo{year}{2022}).
\newblock \bibinfo{title}{Bifidelity surrogate modelling: Showcasing the need
  for new test instances}.
\newblock {\it \bibinfo{journal}{INFORMS Journal on Computing}\/}, .
\bibitem[{Bradford et~al.(2018)Bradford, Schweidtmann, Zhang, Jing \& del
  Rio-Chanona}]{bradford2018dynamic}
\bibinfo{author}{Bradford, E.}, \bibinfo{author}{Schweidtmann, A.~M.},
  \bibinfo{author}{Zhang, D.}, \bibinfo{author}{Jing, K.}, \&
  \bibinfo{author}{del Rio-Chanona, E.~A.} (\bibinfo{year}{2018}).
\newblock \bibinfo{title}{Dynamic modeling and optimization of sustainable
  algal production with uncertainty using multivariate gaussian processes}.
\newblock {\it \bibinfo{journal}{Computers \& Chemical Engineering}\/},  {\it
  \bibinfo{volume}{118}\/}, \bibinfo{pages}{143--158}.
\bibitem[{Castan et~al.(2018)Castan, Schulz, Wenger \& Fischer}]{CASTAN2018131}
\bibinfo{author}{Castan, A.}, \bibinfo{author}{Schulz, P.},
  \bibinfo{author}{Wenger, T.}, \& \bibinfo{author}{Fischer, S.}
  (\bibinfo{year}{2018}).
\newblock \bibinfo{title}{Cell line development}.
\newblock In \bibinfo{editor}{G.~Jagschies}, \bibinfo{editor}{E.~Lindskog},
  \bibinfo{editor}{K.~Lacki}, \& \bibinfo{editor}{P.~Galliher} (Eds.), {\it
  \bibinfo{booktitle}{Biopharmaceutical Processing}\/} (pp.
  \bibinfo{pages}{131--146}).
\newblock \bibinfo{publisher}{Elsevier}.
\bibitem[{Cutajar et~al.(2019)Cutajar, Pullin, Damianou, Lawrence \&
  Gonz{\'a}lez}]{cutajar2019deep}
\bibinfo{author}{Cutajar, K.}, \bibinfo{author}{Pullin, M.},
  \bibinfo{author}{Damianou, A.}, \bibinfo{author}{Lawrence, N.}, \&
  \bibinfo{author}{Gonz{\'a}lez, J.} (\bibinfo{year}{2019}).
\newblock \bibinfo{title}{Deep gaussian processes for multi-fidelity modeling}.
\newblock {\it \bibinfo{journal}{arXiv preprint arXiv:1903.07320}\/}, .
\bibitem[{Damianou \& Lawrence(2012)}]{Damianou2012}
\bibinfo{author}{Damianou, A.~C.}, \& \bibinfo{author}{Lawrence, N.~D.}
  (\bibinfo{year}{2012}).
\newblock \bibinfo{title}{Deep gaussian processes}, .
\newblock \URLprefix \url{http://arxiv.org/abs/1211.0358}.
\bibitem[{Del Rio-Chanona et~al.(2019)Del Rio-Chanona, Ahmed, Wagner, Lu, Zhang
  \& Jing}]{del2019comparison}
\bibinfo{author}{Del Rio-Chanona, E.~A.}, \bibinfo{author}{Ahmed, N.~R.},
  \bibinfo{author}{Wagner, J.}, \bibinfo{author}{Lu, Y.},
  \bibinfo{author}{Zhang, D.}, \& \bibinfo{author}{Jing, K.}
  (\bibinfo{year}{2019}).
\newblock \bibinfo{title}{Comparison of physics-based and data-driven modelling
  techniques for dynamic optimisation of fed-batch bioprocesses}.
\newblock {\it \bibinfo{journal}{Biotechnology and Bioengineering}\/},  {\it
  \bibinfo{volume}{116}\/}, \bibinfo{pages}{2971--2982}.
\bibitem[{Deringer et~al.(2021)Deringer, Bart{\'o}k, Bernstein, Wilkins,
  Ceriotti \& Cs{\'a}nyi}]{deringer2021gaussian}
\bibinfo{author}{Deringer, V.~L.}, \bibinfo{author}{Bart{\'o}k, A.~P.},
  \bibinfo{author}{Bernstein, N.}, \bibinfo{author}{Wilkins, D.~M.},
  \bibinfo{author}{Ceriotti, M.}, \& \bibinfo{author}{Cs{\'a}nyi, G.}
  (\bibinfo{year}{2021}).
\newblock \bibinfo{title}{Gaussian process regression for materials and
  molecules}.
\newblock {\it \bibinfo{journal}{Chemical Reviews}\/},  {\it
  \bibinfo{volume}{121}\/}, \bibinfo{pages}{10073--10141}.
\bibitem[{Duran-Villalobos et~al.(2020)Duran-Villalobos, Goldrick \&
  Lennox}]{duran2020multivariate}
\bibinfo{author}{Duran-Villalobos, C.~A.}, \bibinfo{author}{Goldrick, S.}, \&
  \bibinfo{author}{Lennox, B.} (\bibinfo{year}{2020}).
\newblock \bibinfo{title}{Multivariate statistical process control of an
  industrial-scale fed-batch simulator}.
\newblock {\it \bibinfo{journal}{Computers \& Chemical Engineering}\/},  {\it
  \bibinfo{volume}{132}\/}, \bibinfo{pages}{106620}.
\bibitem[{Fayyad \& Irani(1993)}]{fayyad1993multi}
\bibinfo{author}{Fayyad, U.~M.}, \& \bibinfo{author}{Irani, K.~B.}
  (\bibinfo{year}{1993}).
\newblock \bibinfo{title}{Multi-interval discretization of continuous-valued
  attributes for classification learning}.
\newblock In {\it \bibinfo{booktitle}{Proceedings of the Thirteenth
  International Joint Conference on Artificial Intelligence}\/} (pp.
  \bibinfo{pages}{1022--1029}).
\bibitem[{Frank et~al.(2019)Frank, Dalenogare \& Ayala}]{frank2019industry}
\bibinfo{author}{Frank, A.~G.}, \bibinfo{author}{Dalenogare, L.~S.}, \&
  \bibinfo{author}{Ayala, N.~F.} (\bibinfo{year}{2019}).
\newblock \bibinfo{title}{Industry 4.0 technologies: Implementation patterns in
  manufacturing companies}.
\newblock {\it \bibinfo{journal}{International Journal of Production
  Economics}\/},  {\it \bibinfo{volume}{210}\/}, \bibinfo{pages}{15--26}.
\bibitem[{Fraser(2001)}]{fraser2001biological}
\bibinfo{author}{Fraser, C.~G.} (\bibinfo{year}{2001}).
\newblock {\it \bibinfo{title}{Biological variation: from principles to
  practice}\/}.
\newblock \bibinfo{publisher}{Amer. Assoc. for Clinical Chemistry}.
\bibitem[{Gangadharan et~al.(2021)Gangadharan, Sewell, Turner, Field, Cheeks,
  Oliver, Slater \& Dikicioglu}]{gangadharan2021data}
\bibinfo{author}{Gangadharan, N.}, \bibinfo{author}{Sewell, D.},
  \bibinfo{author}{Turner, R.}, \bibinfo{author}{Field, R.},
  \bibinfo{author}{Cheeks, M.}, \bibinfo{author}{Oliver, S.~G.},
  \bibinfo{author}{Slater, N.~K.}, \& \bibinfo{author}{Dikicioglu, D.}
  (\bibinfo{year}{2021}).
\newblock \bibinfo{title}{Data intelligence for process performance prediction
  in biologics manufacturing}.
\newblock {\it \bibinfo{journal}{Computers \& Chemical Engineering}\/},  {\it
  \bibinfo{volume}{146}\/}, \bibinfo{pages}{107226}.
\bibitem[{Garc{\'\i}a-Camacho et~al.(2016)Garc{\'\i}a-Camacho,
  L{\'o}pez-Rosales, S{\'a}nchez-Mir{\'o}n, Belarbi, Chisti \&
  Molina-Grima}]{garcia2016artificial}
\bibinfo{author}{Garc{\'\i}a-Camacho, F.}, \bibinfo{author}{L{\'o}pez-Rosales,
  L.}, \bibinfo{author}{S{\'a}nchez-Mir{\'o}n, A.}, \bibinfo{author}{Belarbi,
  E.}, \bibinfo{author}{Chisti, Y.}, \& \bibinfo{author}{Molina-Grima, E.}
  (\bibinfo{year}{2016}).
\newblock \bibinfo{title}{Artificial neural network modeling for predicting the
  growth of the microalga karlodinium veneficum}.
\newblock {\it \bibinfo{journal}{Algal Research}\/},  {\it
  \bibinfo{volume}{14}\/}, \bibinfo{pages}{58--64}.
\bibitem[{Gargalo et~al.(2020{\natexlab{a}})Gargalo, Heras, Jones, Udugama,
  Mansouri, Kr{\"u}hne \& Gernaey}]{gargalo2020towards2}
\bibinfo{author}{Gargalo, C.~L.}, \bibinfo{author}{Heras, S. C. d.~l.},
  \bibinfo{author}{Jones, M.~N.}, \bibinfo{author}{Udugama, I.},
  \bibinfo{author}{Mansouri, S.~S.}, \bibinfo{author}{Kr{\"u}hne, U.}, \&
  \bibinfo{author}{Gernaey, K.~V.} (\bibinfo{year}{2020}{\natexlab{a}}).
\newblock \bibinfo{title}{Towards the development of digital twins for the
  bio-manufacturing industry}.
\newblock {\it \bibinfo{journal}{Digital Twins}\/},  (pp.
  \bibinfo{pages}{1--34}).
\bibitem[{Gargalo et~al.(2020{\natexlab{b}})Gargalo, Udugama, Pontius, Lopez,
  Nielsen, Hasanzadeh, Mansouri, Bayer, Junicke \&
  Gernaey}]{gargalo2020towards}
\bibinfo{author}{Gargalo, C.~L.}, \bibinfo{author}{Udugama, I.},
  \bibinfo{author}{Pontius, K.}, \bibinfo{author}{Lopez, P.~C.},
  \bibinfo{author}{Nielsen, R.~F.}, \bibinfo{author}{Hasanzadeh, A.},
  \bibinfo{author}{Mansouri, S.~S.}, \bibinfo{author}{Bayer, C.},
  \bibinfo{author}{Junicke, H.}, \& \bibinfo{author}{Gernaey, K.~V.}
  (\bibinfo{year}{2020}{\natexlab{b}}).
\newblock \bibinfo{title}{Towards smart biomanufacturing: a perspective on
  recent developments in industrial measurement and monitoring technologies for
  bio-based production processes}.
\newblock {\it \bibinfo{journal}{Journal of Industrial Microbiology \&
  Biotechnology: Official Journal of the Society for Industrial Microbiology
  and Biotechnology}\/},  {\it \bibinfo{volume}{47}\/},
  \bibinfo{pages}{947--964}.
\bibitem[{Glassey et~al.(1994)Glassey, Montague, Ward \&
  Kara}]{glassey1994artificial}
\bibinfo{author}{Glassey, J.}, \bibinfo{author}{Montague, G.},
  \bibinfo{author}{Ward, A.}, \& \bibinfo{author}{Kara, B.}
  (\bibinfo{year}{1994}).
\newblock \bibinfo{title}{Artificial neural network based experimental design
  procedure for enhancing fermentation development}.
\newblock {\it \bibinfo{journal}{Biotechnology and Bioengineering}\/},  {\it
  \bibinfo{volume}{44}\/}, \bibinfo{pages}{397--405}.
\bibitem[{Graefe et~al.(1999)Graefe, Bogaerts, Castillo, Cherlet, W{\'e}renne,
  Marenbach \& Hanus}]{graefe1999new}
\bibinfo{author}{Graefe, J.}, \bibinfo{author}{Bogaerts, P.},
  \bibinfo{author}{Castillo, J.}, \bibinfo{author}{Cherlet, M.},
  \bibinfo{author}{W{\'e}renne, J.}, \bibinfo{author}{Marenbach, P.}, \&
  \bibinfo{author}{Hanus, R.} (\bibinfo{year}{1999}).
\newblock \bibinfo{title}{A new training method for hybrid models of
  bioprocesses}.
\newblock {\it \bibinfo{journal}{Bioprocess Engineering}\/},  {\it
  \bibinfo{volume}{21}\/}, \bibinfo{pages}{423--429}.
\bibitem[{Gratiet \& Garnier(2014)}]{Gratiet2014}
\bibinfo{author}{Gratiet, L.~L.}, \& \bibinfo{author}{Garnier, J.}
  (\bibinfo{year}{2014}).
\newblock \bibinfo{title}{Recursive co-kriging model for design of computer
  experiments with multiple levels of fidelity}.
\newblock {\it \bibinfo{journal}{International Journal for Uncertainty
  Quantification}\/},  {\it \bibinfo{volume}{4}\/}.
  \DOIprefix\doi{10.1615/int.j.uncertaintyquantification.2014006914}.
\bibitem[{Hutter et~al.(2021)Hutter, von Stosch, Bournazou \&
  Butt{\'e}}]{hutter2021knowledge}
\bibinfo{author}{Hutter, C.}, \bibinfo{author}{von Stosch, M.},
  \bibinfo{author}{Bournazou, M. N.~C.}, \& \bibinfo{author}{Butt{\'e}, A.}
  (\bibinfo{year}{2021}).
\newblock \bibinfo{title}{Knowledge transfer across cell lines using hybrid
  gaussian process models with entity embedding vectors}.
\newblock {\it \bibinfo{journal}{Biotechnology and Bioengineering}\/},  {\it
  \bibinfo{volume}{118}\/}, \bibinfo{pages}{4389--4401}.
\bibitem[{Jang \& Barford(2000)}]{jang2000unstructured}
\bibinfo{author}{Jang, J.~D.}, \& \bibinfo{author}{Barford, J.~P.}
  (\bibinfo{year}{2000}).
\newblock \bibinfo{title}{An unstructured kinetic model of macromolecular
  metabolism in batch and fed-batch cultures of hybridoma cells producing
  monoclonal antibody}.
\newblock {\it \bibinfo{journal}{Biochemical Engineering Journal}\/},  {\it
  \bibinfo{volume}{4}\/}, \bibinfo{pages}{153--168}.
\bibitem[{Kennedy \& O'Hagan(2000)}]{Kennedy2000}
\bibinfo{author}{Kennedy, M.~C.}, \& \bibinfo{author}{O'Hagan, A.}
  (\bibinfo{year}{2000}).
\newblock \bibinfo{title}{Predicting the output from a complex computer code
  when fast approximations are available}.
\newblock {\it \bibinfo{journal}{Biometrika}\/},  {\it \bibinfo{volume}{87}\/}.
  \DOIprefix\doi{10.1093/biomet/87.1.1}.
\bibitem[{Kyriakopoulos et~al.(2018)Kyriakopoulos, Ang, Lakshmanan, Huang,
  Yoon, Gunawan \& Lee}]{kyriakopoulos2018kinetic}
\bibinfo{author}{Kyriakopoulos, S.}, \bibinfo{author}{Ang, K.~S.},
  \bibinfo{author}{Lakshmanan, M.}, \bibinfo{author}{Huang, Z.},
  \bibinfo{author}{Yoon, S.}, \bibinfo{author}{Gunawan, R.}, \&
  \bibinfo{author}{Lee, D.-Y.} (\bibinfo{year}{2018}).
\newblock \bibinfo{title}{Kinetic modeling of mammalian cell culture
  bioprocessing: the quest to advance biomanufacturing}.
\newblock {\it \bibinfo{journal}{Biotechnology Journal}\/},  {\it
  \bibinfo{volume}{13}\/}, \bibinfo{pages}{1700229}.
\bibitem[{Lasi et~al.(2014)Lasi, Fettke, Kemper, Feld \&
  Hoffmann}]{lasi2014industry}
\bibinfo{author}{Lasi, H.}, \bibinfo{author}{Fettke, P.},
  \bibinfo{author}{Kemper, H.-G.}, \bibinfo{author}{Feld, T.}, \&
  \bibinfo{author}{Hoffmann, M.} (\bibinfo{year}{2014}).
\newblock \bibinfo{title}{Industry 4.0}.
\newblock {\it \bibinfo{journal}{Business \& Information Systems
  Engineering}\/},  {\it \bibinfo{volume}{6}\/}, \bibinfo{pages}{239--242}.
\bibitem[{Lindskog(2018)}]{lindskog2018upstream}
\bibinfo{author}{Lindskog, E.~K.} (\bibinfo{year}{2018}).
\newblock \bibinfo{title}{The upstream process: principal modes of operation}.
\newblock In {\it \bibinfo{booktitle}{Biopharmaceutical Processing}\/} (pp.
  \bibinfo{pages}{625--635}).
\newblock \bibinfo{publisher}{Elsevier}.
\bibitem[{Lindskog et~al.(2018)Lindskog, Fischer, Wenger \&
  Schulz}]{lindskog2018host}
\bibinfo{author}{Lindskog, E.~K.}, \bibinfo{author}{Fischer, S.},
  \bibinfo{author}{Wenger, T.}, \& \bibinfo{author}{Schulz, P.}
  (\bibinfo{year}{2018}).
\newblock \bibinfo{title}{Host cells}.
\newblock In {\it \bibinfo{booktitle}{Biopharmaceutical Processing}\/} (pp.
  \bibinfo{pages}{111--130}).
\newblock \bibinfo{publisher}{Elsevier}.
\bibitem[{Mercier et~al.(2013)Mercier, Diepenbroek, Dalm, Wijffels \&
  Streefland}]{mercier2013multivariate}
\bibinfo{author}{Mercier, S.~M.}, \bibinfo{author}{Diepenbroek, B.},
  \bibinfo{author}{Dalm, M.~C.}, \bibinfo{author}{Wijffels, R.~H.}, \&
  \bibinfo{author}{Streefland, M.} (\bibinfo{year}{2013}).
\newblock \bibinfo{title}{Multivariate data analysis as a pat tool for early
  bioprocess development data}.
\newblock {\it \bibinfo{journal}{Journal of Biotechnology}\/},  {\it
  \bibinfo{volume}{167}\/}, \bibinfo{pages}{262--270}.
\bibitem[{Merkelbach et~al.(2022)Merkelbach, Schweidtmann, M{\"u}ller,
  Schwoebel, Mhamdi, Mitsos, Schuppert, Mrziglod \&
  Schneckener}]{merkelbach2022hybridml}
\bibinfo{author}{Merkelbach, K.}, \bibinfo{author}{Schweidtmann, A.~M.},
  \bibinfo{author}{M{\"u}ller, Y.}, \bibinfo{author}{Schwoebel, P.},
  \bibinfo{author}{Mhamdi, A.}, \bibinfo{author}{Mitsos, A.},
  \bibinfo{author}{Schuppert, A.}, \bibinfo{author}{Mrziglod, T.}, \&
  \bibinfo{author}{Schneckener, S.} (\bibinfo{year}{2022}).
\newblock \bibinfo{title}{{HybridML}: Open source platform for hybrid
  modeling}.
\newblock {\it \bibinfo{journal}{Computers \& Chemical Engineering}\/},  {\it
  \bibinfo{volume}{160}\/}, \bibinfo{pages}{107736}.
\bibitem[{Mu{\~n}oz et~al.(2015)Mu{\~n}oz, Sun, Kirley \&
  Halgamuge}]{munoz2015algorithm}
\bibinfo{author}{Mu{\~n}oz, M.~A.}, \bibinfo{author}{Sun, Y.},
  \bibinfo{author}{Kirley, M.}, \& \bibinfo{author}{Halgamuge, S.~K.}
  (\bibinfo{year}{2015}).
\newblock \bibinfo{title}{Algorithm selection for black-box continuous
  optimization problems: A survey on methods and challenges}.
\newblock {\it \bibinfo{journal}{Information Sciences}\/},  {\it
  \bibinfo{volume}{317}\/}, \bibinfo{pages}{224--245}.
\bibitem[{Narayanan et~al.(2019)Narayanan, Sokolov, Morbidelli \&
  Butt{\'e}}]{narayanan2019new}
\bibinfo{author}{Narayanan, H.}, \bibinfo{author}{Sokolov, M.},
  \bibinfo{author}{Morbidelli, M.}, \& \bibinfo{author}{Butt{\'e}, A.}
  (\bibinfo{year}{2019}).
\newblock \bibinfo{title}{A new generation of predictive models: The added
  value of hybrid models for manufacturing processes of therapeutic proteins}.
\newblock {\it \bibinfo{journal}{Biotechnology and Bioengineering}\/},  {\it
  \bibinfo{volume}{116}\/}, \bibinfo{pages}{2540--2549}.
\bibitem[{O'Hagan(1978)}]{o1978curve}
\bibinfo{author}{O'Hagan, A.} (\bibinfo{year}{1978}).
\newblock \bibinfo{title}{Curve fitting and optimal design for prediction}.
\newblock {\it \bibinfo{journal}{Journal of the Royal Statistical Society:
  Series B (Methodological)}\/},  {\it \bibinfo{volume}{40}\/},
  \bibinfo{pages}{1--24}.
\bibitem[{Papathanasiou \& Kontoravdi(2020)}]{papathanasiou2020engineering}
\bibinfo{author}{Papathanasiou, M.~M.}, \& \bibinfo{author}{Kontoravdi, C.}
  (\bibinfo{year}{2020}).
\newblock \bibinfo{title}{Engineering challenges in therapeutic protein product
  and process design}.
\newblock {\it \bibinfo{journal}{Current Opinion in Chemical Engineering}\/},
  {\it \bibinfo{volume}{27}\/}, \bibinfo{pages}{81--88}.
\bibitem[{Peherstorfer et~al.(2018)Peherstorfer, Willcox \&
  Gunzburger}]{Peherstorfer2018}
\bibinfo{author}{Peherstorfer, B.}, \bibinfo{author}{Willcox, K.}, \&
  \bibinfo{author}{Gunzburger, M.} (\bibinfo{year}{2018}).
\newblock \bibinfo{title}{Survey of multifidelity methods in uncertainty
  propagation, inference, and optimization}.
\newblock \DOIprefix\doi{10.1137/16M1082469}.
\bibitem[{Peng et~al.(2005)Peng, Long \& Ding}]{peng2005feature}
\bibinfo{author}{Peng, H.}, \bibinfo{author}{Long, F.}, \&
  \bibinfo{author}{Ding, C.} (\bibinfo{year}{2005}).
\newblock \bibinfo{title}{Feature selection based on mutual information
  criteria of max-dependency, max-relevance, and min-redundancy}.
\newblock {\it \bibinfo{journal}{IEEE Transactions on Pattern Analysis and
  Machine Intelligence}\/},  {\it \bibinfo{volume}{27}\/},
  \bibinfo{pages}{1226--1238}.
\bibitem[{Perdikaris et~al.(2017)Perdikaris, Raissi, Damianou, Lawrence \&
  Karniadakis}]{Perdikaris2017}
\bibinfo{author}{Perdikaris, P.}, \bibinfo{author}{Raissi, M.},
  \bibinfo{author}{Damianou, A.}, \bibinfo{author}{Lawrence, N.~D.}, \&
  \bibinfo{author}{Karniadakis, G.~E.} (\bibinfo{year}{2017}).
\newblock \bibinfo{title}{Nonlinear information fusion algorithms for
  data-efficient multi-fidelity modelling}.
\newblock {\it \bibinfo{journal}{Proceedings of the Royal Society A:
  Mathematical, Physical and Engineering Sciences}\/},  {\it
  \bibinfo{volume}{473}\/}. \DOIprefix\doi{10.1098/rspa.2016.0751}.
\bibitem[{Petsagkourakis \& Galvanin(2021)}]{petsagkourakis2021safe}
\bibinfo{author}{Petsagkourakis, P.}, \& \bibinfo{author}{Galvanin, F.}
  (\bibinfo{year}{2021}).
\newblock \bibinfo{title}{Safe model-based design of experiments using gaussian
  processes}.
\newblock {\it \bibinfo{journal}{Computers \& Chemical Engineering}\/},  {\it
  \bibinfo{volume}{151}\/}, \bibinfo{pages}{107339}.
\bibitem[{Petsagkourakis et~al.(2020)Petsagkourakis, Sandoval, Bradford, Zhang
  \& del Rio-Chanona}]{petsagkourakis2020reinforcement}
\bibinfo{author}{Petsagkourakis, P.}, \bibinfo{author}{Sandoval, I.~O.},
  \bibinfo{author}{Bradford, E.}, \bibinfo{author}{Zhang, D.}, \&
  \bibinfo{author}{del Rio-Chanona, E.~A.} (\bibinfo{year}{2020}).
\newblock \bibinfo{title}{Reinforcement learning for batch bioprocess
  optimization}.
\newblock {\it \bibinfo{journal}{Computers \& Chemical Engineering}\/},  {\it
  \bibinfo{volume}{133}\/}, \bibinfo{pages}{106649}.
\bibitem[{Rasmussen \& Williams(2018)}]{Rasmussen2018}
\bibinfo{author}{Rasmussen, C.~E.}, \& \bibinfo{author}{Williams, C. K.~I.}
  (\bibinfo{year}{2018}).
\newblock {\it \bibinfo{title}{Gaussian Processes for Machine Learning}\/}.
\newblock \DOIprefix\doi{10.7551/mitpress/3206.001.0001}.
\bibitem[{Richelle et~al.(2020)Richelle, Lee, Portela, Raley \& von
  Stosch}]{richelle2020analysis}
\bibinfo{author}{Richelle, A.}, \bibinfo{author}{Lee, B.~W.},
  \bibinfo{author}{Portela, R.~M.}, \bibinfo{author}{Raley, J.}, \&
  \bibinfo{author}{von Stosch, M.} (\bibinfo{year}{2020}).
\newblock \bibinfo{title}{Analysis of transformed upstream bioprocess data
  provides insights into biological system variation}.
\newblock {\it \bibinfo{journal}{Biotechnology Journal}\/},  {\it
  \bibinfo{volume}{15}\/}, \bibinfo{pages}{2000113}.
\bibitem[{del Rio-Chanona et~al.(2017{\natexlab{a}})del Rio-Chanona, Ahmed,
  Zhang, Lu \& Jing}]{del2017kinetic}
\bibinfo{author}{del Rio-Chanona, E.~A.}, \bibinfo{author}{Ahmed, N.~r.},
  \bibinfo{author}{Zhang, D.}, \bibinfo{author}{Lu, Y.}, \&
  \bibinfo{author}{Jing, K.} (\bibinfo{year}{2017}{\natexlab{a}}).
\newblock \bibinfo{title}{Kinetic modeling and process analysis for desmodesmus
  sp. lutein photo-production}.
\newblock {\it \bibinfo{journal}{AIChE Journal}\/},  {\it
  \bibinfo{volume}{63}\/}, \bibinfo{pages}{2546--2554}.
\bibitem[{del Rio-Chanona et~al.(2017{\natexlab{b}})del Rio-Chanona, Fiorelli,
  Zhang, Ahmed, Jing \& Shah}]{del2017efficient}
\bibinfo{author}{del Rio-Chanona, E.~A.}, \bibinfo{author}{Fiorelli, F.},
  \bibinfo{author}{Zhang, D.}, \bibinfo{author}{Ahmed, N.~R.},
  \bibinfo{author}{Jing, K.}, \& \bibinfo{author}{Shah, N.}
  (\bibinfo{year}{2017}{\natexlab{b}}).
\newblock \bibinfo{title}{An efficient model construction strategy to simulate
  microalgal lutein photo-production dynamic process}.
\newblock {\it \bibinfo{journal}{Biotechnology and Bioengineering}\/},  {\it
  \bibinfo{volume}{114}\/}, \bibinfo{pages}{2518--2527}.
\bibitem[{del Rio~Chanona et~al.(2021)del Rio~Chanona, Petsagkourakis,
  Bradford, Graciano \& Chachuat}]{del2021real}
\bibinfo{author}{del Rio~Chanona, E.~A.}, \bibinfo{author}{Petsagkourakis, P.},
  \bibinfo{author}{Bradford, E.}, \bibinfo{author}{Graciano, J.~A.}, \&
  \bibinfo{author}{Chachuat, B.} (\bibinfo{year}{2021}).
\newblock \bibinfo{title}{Real-time optimization meets bayesian optimization
  and derivative-free optimization: A tale of modifier adaptation}.
\newblock {\it \bibinfo{journal}{Computers \& Chemical Engineering}\/},  {\it
  \bibinfo{volume}{147}\/}, \bibinfo{pages}{107249}.
\bibitem[{Rogers et~al.(2022)Rogers, Vega-Ramon, Yan, del R{\'\i}o-Chanona,
  Jing \& Zhang}]{rogers2022transfer}
\bibinfo{author}{Rogers, A.~W.}, \bibinfo{author}{Vega-Ramon, F.},
  \bibinfo{author}{Yan, J.}, \bibinfo{author}{del R{\'\i}o-Chanona, E.~A.},
  \bibinfo{author}{Jing, K.}, \& \bibinfo{author}{Zhang, D.}
  (\bibinfo{year}{2022}).
\newblock \bibinfo{title}{A transfer learning approach for predictive modeling
  of bioprocesses using small data}.
\newblock {\it \bibinfo{journal}{Biotechnology and Bioengineering}\/},  {\it
  \bibinfo{volume}{119}\/}, \bibinfo{pages}{411--422}.
\bibitem[{di~Sciascio \& Amicarelli(2008)}]{di2008biomass}
\bibinfo{author}{di~Sciascio, F.}, \& \bibinfo{author}{Amicarelli, A.~N.}
  (\bibinfo{year}{2008}).
\newblock \bibinfo{title}{Biomass estimation in batch biotechnological
  processes by bayesian gaussian process regression}.
\newblock {\it \bibinfo{journal}{Computers \& Chemical Engineering}\/},  {\it
  \bibinfo{volume}{32}\/}, \bibinfo{pages}{3264--3273}.
\bibitem[{Sokolov et~al.(2021)Sokolov, von Stosch, Narayanan, Feidl \&
  Butt{\'e}}]{sokolov2021hybrid}
\bibinfo{author}{Sokolov, M.}, \bibinfo{author}{von Stosch, M.},
  \bibinfo{author}{Narayanan, H.}, \bibinfo{author}{Feidl, F.}, \&
  \bibinfo{author}{Butt{\'e}, A.} (\bibinfo{year}{2021}).
\newblock \bibinfo{title}{Hybrid modeling—a key enabler towards realizing
  digital twins in biopharma?}
\newblock {\it \bibinfo{journal}{Current Opinion in Chemical Engineering}\/},
  {\it \bibinfo{volume}{34}\/}, \bibinfo{pages}{100715}.
\bibitem[{Solle et~al.(2017)Solle, Hitzmann, Herwig, Pereira~Remelhe, Ulonska,
  Wuerth, Prata \& Steckenreiter}]{solle2017between}
\bibinfo{author}{Solle, D.}, \bibinfo{author}{Hitzmann, B.},
  \bibinfo{author}{Herwig, C.}, \bibinfo{author}{Pereira~Remelhe, M.},
  \bibinfo{author}{Ulonska, S.}, \bibinfo{author}{Wuerth, L.},
  \bibinfo{author}{Prata, A.}, \& \bibinfo{author}{Steckenreiter, T.}
  (\bibinfo{year}{2017}).
\newblock \bibinfo{title}{Between the poles of data-driven and mechanistic
  modeling for process operation}.
\newblock {\it \bibinfo{journal}{Chemie Ingenieur Technik}\/},  {\it
  \bibinfo{volume}{89}\/}, \bibinfo{pages}{542--561}.
\bibitem[{Sun et~al.(2020)Sun, Wang, Kirley, Li \& Chan}]{sun2020revisiting}
\bibinfo{author}{Sun, Y.}, \bibinfo{author}{Wang, W.}, \bibinfo{author}{Kirley,
  M.}, \bibinfo{author}{Li, X.}, \& \bibinfo{author}{Chan, J.}
  (\bibinfo{year}{2020}).
\newblock \bibinfo{title}{Revisiting probability distribution assumptions for
  information theoretic feature selection}.
\newblock In {\it \bibinfo{booktitle}{Proceedings of the AAAI Conference on
  Artificial Intelligence}\/} (pp. \bibinfo{pages}{5908--5915}).
\newblock volume~\bibinfo{volume}{34}.
\bibitem[{Tescione et~al.(2015)Tescione, Lambropoulos, Paranandi, Makagiansar
  \& Ryll}]{tescione2015application}
\bibinfo{author}{Tescione, L.}, \bibinfo{author}{Lambropoulos, J.},
  \bibinfo{author}{Paranandi, M.~R.}, \bibinfo{author}{Makagiansar, H.}, \&
  \bibinfo{author}{Ryll, T.} (\bibinfo{year}{2015}).
\newblock \bibinfo{title}{Application of bioreactor design principles and
  multivariate analysis for development of cell culture scale down models}.
\newblock {\it \bibinfo{journal}{Biotechnology and Bioengineering}\/},  {\it
  \bibinfo{volume}{112}\/}, \bibinfo{pages}{84--97}.
\bibitem[{Tsopanoglou \& del Val(2021)}]{tsopanoglou2021moving}
\bibinfo{author}{Tsopanoglou, A.}, \& \bibinfo{author}{del Val, I.~J.}
  (\bibinfo{year}{2021}).
\newblock \bibinfo{title}{Moving towards an era of hybrid modelling: advantages
  and challenges of coupling mechanistic and data-driven models for upstream
  pharmaceutical bioprocesses}.
\newblock {\it \bibinfo{journal}{Current Opinion in Chemical Engineering}\/},
  {\it \bibinfo{volume}{32}\/}, \bibinfo{pages}{100691}.
\bibitem[{Udugama et~al.(2020)Udugama, Gargalo, Yamashita, Taube, Palazoglu,
  Young, Gernaey, Kulahci \& Bayer}]{udugama2020role}
\bibinfo{author}{Udugama, I.~A.}, \bibinfo{author}{Gargalo, C.~L.},
  \bibinfo{author}{Yamashita, Y.}, \bibinfo{author}{Taube, M.~A.},
  \bibinfo{author}{Palazoglu, A.}, \bibinfo{author}{Young, B.~R.},
  \bibinfo{author}{Gernaey, K.~V.}, \bibinfo{author}{Kulahci, M.}, \&
  \bibinfo{author}{Bayer, C.} (\bibinfo{year}{2020}).
\newblock \bibinfo{title}{The role of big data in industrial (bio) chemical
  process operations}.
\newblock {\it \bibinfo{journal}{Industrial \& Engineering Chemistry
  Research}\/},  {\it \bibinfo{volume}{59}\/}, \bibinfo{pages}{15283--15297}.
\bibitem[{Xia et~al.(2015)Xia, Wang, Lin, Wang, Chu, Zhuang \&
  Zhang}]{xia2015advances}
\bibinfo{author}{Xia, J.}, \bibinfo{author}{Wang, G.}, \bibinfo{author}{Lin,
  J.}, \bibinfo{author}{Wang, Y.}, \bibinfo{author}{Chu, J.},
  \bibinfo{author}{Zhuang, Y.}, \& \bibinfo{author}{Zhang, S.}
  (\bibinfo{year}{2015}).
\newblock \bibinfo{title}{Advances and practices of bioprocess scale-up}.
\newblock {\it \bibinfo{journal}{Bioreactor Engineering Research and Industrial
  Applications II}\/},  (pp. \bibinfo{pages}{137--151}).
\bibitem[{Zhang et~al.(2013)Zhang, Hunter \& Zhou}]{ZHANG2013127}
\bibinfo{author}{Zhang, J.}, \bibinfo{author}{Hunter, A.}, \&
  \bibinfo{author}{Zhou, Y.} (\bibinfo{year}{2013}).
\newblock \bibinfo{title}{A logic-reasoning based system to harness bioprocess
  experimental data and knowledge for design}.
\newblock {\it \bibinfo{journal}{Biochemical Engineering Journal}\/},  {\it
  \bibinfo{volume}{74}\/}, \bibinfo{pages}{127--135}.

\end{thebibliography}

\end{document}